\documentclass[11pt,a4paper]{article}
\usepackage[hyperref]{acl2020}
\usepackage{latexsym}
\usepackage{times}
\usepackage{subcaption}
\usepackage{graphicx}
\usepackage{comment}
\usepackage{color}
\usepackage{booktabs}
\usepackage{amsmath}
\usepackage{amssymb}
\usepackage{dblfloatfix}
\usepackage{pbox}
\usepackage{array}
\usepackage{url}

\usepackage{microtype}

\aclfinalcopy 

\title{Exploration of Gender Differences in COVID-19 Discourse on Reddit}

\author{
Jai Aggarwal \hspace{2.8cm}
Ella Rabinovich \hspace{2.8cm}
Suzanne Stevenson \vspace{0.2cm} \\
Department of Computer Science, University of Toronto \vspace{0.1cm} \\
\texttt{\{jai,ella,suzanne\}@cs.toronto.edu}
}

\date{}

\begin{document}
\maketitle
\begin{abstract}
Decades of research on differences in the language of men and women have established postulates about preferences in lexical, topical, and emotional expression between the two genders, along with their sociological underpinnings. Using a novel dataset of male and female linguistic productions collected from the Reddit discussion platform, we further confirm existing assumptions about gender-linked affective distinctions, and demonstrate that these distinctions are amplified in social media postings involving
emotionally-charged discourse related to COVID-19. Our analysis also confirms
considerable differences in topical preferences between male and female authors in spontaneous pandemic-related discussions.
\end{abstract}

\section{Introduction}

Research on gender differences in language has a long history spanning psychology, gender studies, sociolinguistics, and, more recently, computational linguistics. A considerable body of linguistic studies highlights the differences between the language of men and women in topical, lexical, and syntactic aspects \citep{lakoff1973language, labov1990intersection}, and such differences have proven to be accurately detectable by automatic classification tools \citep{koppel2002automatically,schler2006effects, schwartz2013personality}. Here, we study the differences in male (M) and female (F) language in discussions of COVID-19\footnote{We refer to COVID-19 by `COVID' hereafter.} on the Reddit\footnote{\url{https://www.reddit.com/}} discussion platform. Responses to the virus on social media have been heavily emotionally-charged, accompanied by feelings of anxiety, grief, and fear, and have discussed far-ranging concerns regarding personal and public health, the economy, and social aspects of life. In this work, we explore how established emotional and topical cross-gender differences are carried over into this pandemic-related discourse. Insights regrading these distinctions will advance our understanding of gender-linked linguistic traits, and may further help to inform public policy and communications around the pandemic.

Research has considered the emotional content of social media on the topic of the COVID pandemic \citep[e.g.,][]{LwinEtAl2020, StellaEtAl2020}, but little work has looked specifically at the impact of gender on affective expression \citep{vandervegt2020women}. Gender-linked linguistic distinctions across emotional dimensions have been a subject of prolific research \citep{burriss2007psychophysiological, hoffman2008empathy, thelwall2010data}, with findings suggesting that women are more likely than men to express positive emotions, while men exhibit higher tendency to dominance, engagement, and control (although see \citet{park2016women} for an alternative finding). \citet{vandervegt2020women} compared the self-reported emotional state of male vs.\ female crowdsourced workers who contributed to the Real World Worry Dataset \citep[RWWD,][]{RWWD2020}, in which they were also asked to write about their feelings around COVID.  However, because \citet{vandervegt2020women} restricted the affective analysis to the workers’ emotional ratings, it remains an open question regarding whether, and how, the natural linguistic productions of males and females about COVID will exhibit detectably different patterns of emotion.

Topical analysis of social media during the pandemic has also been a focus of recent work \citep[e.g.,][]{liu_health_2020, abd-alrazaq_top_2020}, again with few studies devoted to gender differences \citep{thelwall_covid-19_2020, vandervegt2020women}. Much prior work has found distinctions in topical preferences in spontaneous productions of the two genders \citep[e.g.,][]{mulac2001empirical, mulac2006gender, newman2008gender}, showing that men were more likely to discuss money- and occupation-related topics, focused on objects and impersonal matters, while women preferred discussion on family and social life, topics related to psychological and social processes.  In the recent context, \citet{thelwall_covid-19_2020} found these observations persisted in COVID-19 tweets, with a male focus on sports and politics, and female focus on family and caring.  In the prompted texts of the RWWD, \citet{vandervegt2020women} also found the expected M vs.\ F topical differences, with men talking more about the international impact of the pandemic, as well as governmental policy, and women more commonly discussing social aspects -- family, friends, and solidarity.  Moreover, \citet{vandervegt2020women} further found differences between the elicited short (tweet-sized) and longer essays, revealing the 
impact of the goal and size of the text on such analyses.  Again, an open question remains concerning the topical distinctions between M and F authors in spontaneous productions without artificial restrictions on length. 

Here, we aim to address the above gaps in the literature, by performing a comprehensive analysis of the similarities and differences between male and female language collected from the Reddit discussion platform.  Our main corpus is a large collection of spontaneous COVID-related utterances by (self-reported) M and F authors.  Importantly, we also collect productions on a wide variety of topics by the same set of authors as a `baseline' dataset.  First, using a multidimensional affective framework from psychology \citep{bradley1994measuring}, we draw on a recently-released dataset of human affective ratings of words \citet{mohammad2018obtaining} to support the emotional assessment of male and female posts in our datasets.  Through this approach, we corroborate existing assumptions on differences in the emotional aspects of linguistic productions of men and women in the COVID corpus.  Moreover, our use of a baseline dataset enables us to further show that these distinctions are amplified in the emotionally-intensive setting of COVID discussions compared to productions on other topics. Second, we take a topic modeling approach to demonstrate detectable distinctions in the range of topics discussed by the two genders in our COVID corpus, reinforcing (to some extent) assumptions on gender-related topical preferences, in this natural discourse in an emotionally-charged context.\footnote{All data and code is available at \url{https://github.com/ellarabi/covid19-demography}.}

\section{Datasets}

As noted above, our goal is to analyze emotions and topics in spontaneous utterances that are relatively unconstrained by length.  To that end,
our main dataset comprises a large collection of spontaneous, COVID-related English utterances by male and female authors from the Reddit discussion platforms. As of May 2020, Reddit had
over $430$M active users, $1.2$M topical threads (subreddits), and over $70$\% of its user base coming from English-speaking countries. Subreddits often encourage their subscribers to specify a meta-property (called a `flair', a textual tag), projecting a small glimpse about themselves (e.g., political association, country of origin, age), thereby customizing their presence within a subreddit. 

\begin{figure}[hbt]
\centering
\includegraphics[width=7cm]{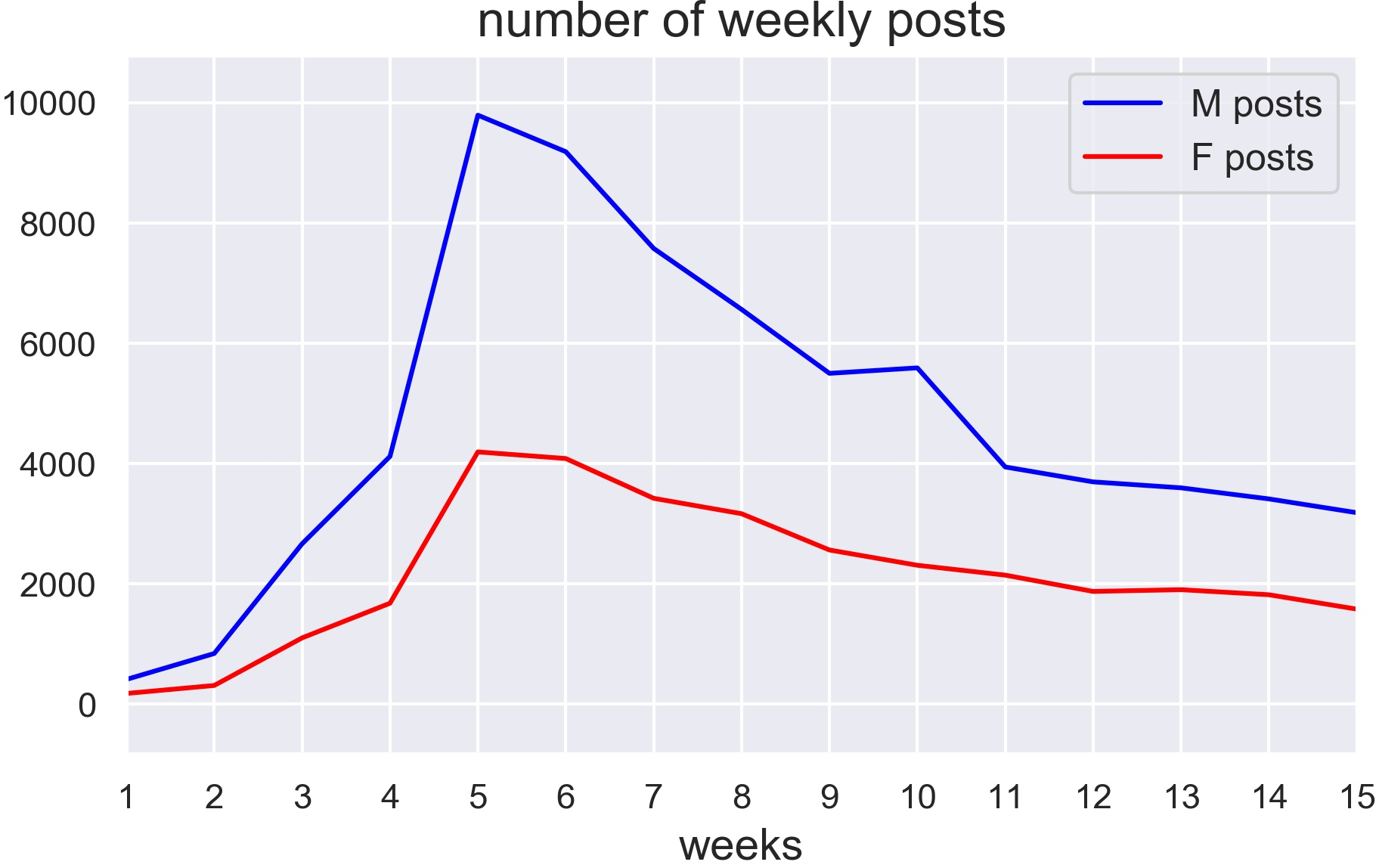}
\caption{Weekly COVID-related posts by gender.}
\label{fig:weekly-counts}
\end{figure}

We identified a set of subreddits, such as `r/askmen' and `r/askwomen', where authors commonly self-report their gender, and extracted a set of unique user-ids of authors who specified male or female gender as a flair.\footnote{Although gender can be viewed as a continuum rather than binary, we limit this study to the two most prominent gender markers in our corpus: male and female.} This process yielded the user-ids for $10,421$ males and $5,630$ females (as self-reported).  Using this extracted set of ids,
we collected COVID-related submissions and comments\footnote{For convenience, we refer to both initial submissions and comments to submissions as `posts' hereafter.} from across the Reddit discussion platform for a period of 15 weeks, from February 1st through June 1st.  COVID-related posts were identified as those containing one or more of a set of predefined keywords: `covid', `covid-19', `covid19', `corona', `coronavirus', `the virus', `pandemic'. 
This process resulted in over $70$K male and $35$K female posts spanning $7,583$ topical threads; the male subcorpus contains $5.3$M tokens and the female subcorpus $2.8$M tokens.
Figure~\ref{fig:weekly-counts} presents the weekly amount of COVID-related posts in the combined corpus, showing a peak in early-mid March (weeks $5$--$6$).

Aiming at a comparative analysis between virus-related and `neutral' (baseline) linguistic productions by men and women, we collected an additional dataset comprising a randomly sampled $10$K posts per week by the same set of authors, totalling $150$K posts for each gender. The baseline dataset contains $6.8$M tokens in the male subcorpus and $5.3$M tokens in the female subcorpus.

We use our COVID and baseline datasets for analysis of emotional differences as well as topical preferences in spontaneous productions by male and female authors on Reddit.  The ample size of the corpora facilitates analysis of distinctions in these two aspects between the two genders in their discourse on the pandemic, and as compared to non-COVID discussion.

\section{Analysis of Emotional Dimensions}

\subsection{Methods}

\begin{table*}
\resizebox{\textwidth}{!}{
\begin{tabular}{l|rr|rr|r||rr|rr|r}
\multicolumn{1}{c}{} & \multicolumn{5}{c||}{COVID-related posts} & \multicolumn{5}{c}{Baseline posts} \\
& mean(M) & std(M) & mean(F) & std(F) & eff. size & mean(M) & std(M) & mean(F) & std(F) & eff. size \\ \hline
V & 0.375 & 0.12 & \textbf{0.388} & 0.11 &  -0.120 & 0.453 & 0.14 & \textbf{0.459} & 0.14 & -0.043 \\
A & \textbf{0.579} & 0.09 & 0.567 & 0.08 & 0.144   & \textbf{0.570} & 0.10 & 0.559 & 0.09 & 0.109 \\
D & \textbf{0.490} & 0.08 & 0.476 & 0.07 & 0.183   & \textbf{0.486} & 0.09 & 0.469 & 0.09 & 0.185 \\
\end{tabular}
}
\caption{\label{tbl:vad-values} Means of M and F posts for each affective dimension, and effect size of differences within each corpus. All differences significant at p\textless$0.001$. Highest mean score for each of V, A, D, in COVID and baseline, is boldfaced.}
\end{table*}

\begin{figure*}[ht!]
\begin{subfigure}[t]{0.1\textwidth}
    \includegraphics[scale=0.4]{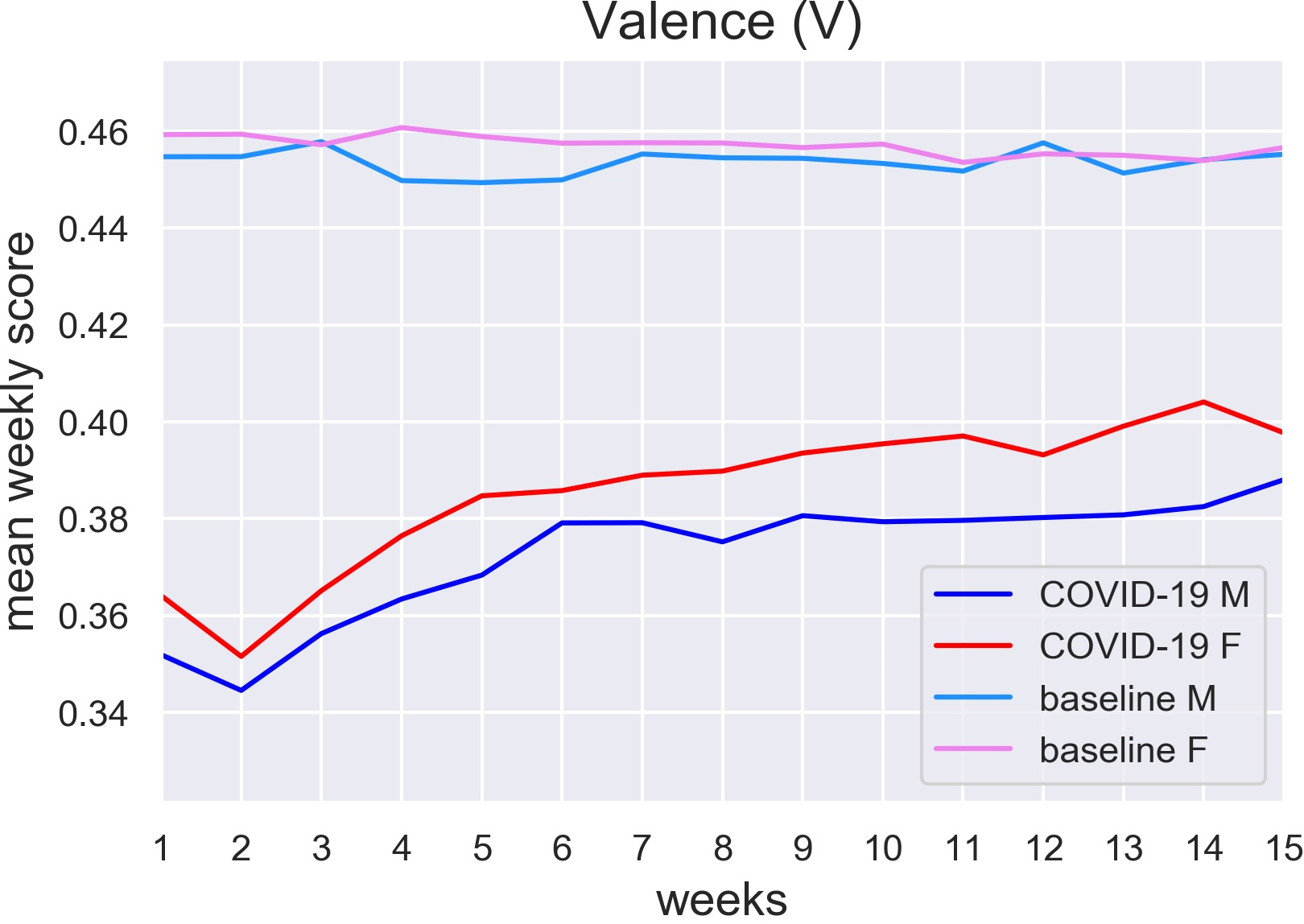}
\end{subfigure}
\qquad \qquad \quad \qquad \qquad \quad
\begin{subfigure}[t]{0.1\textwidth}
    \includegraphics[scale=0.4]{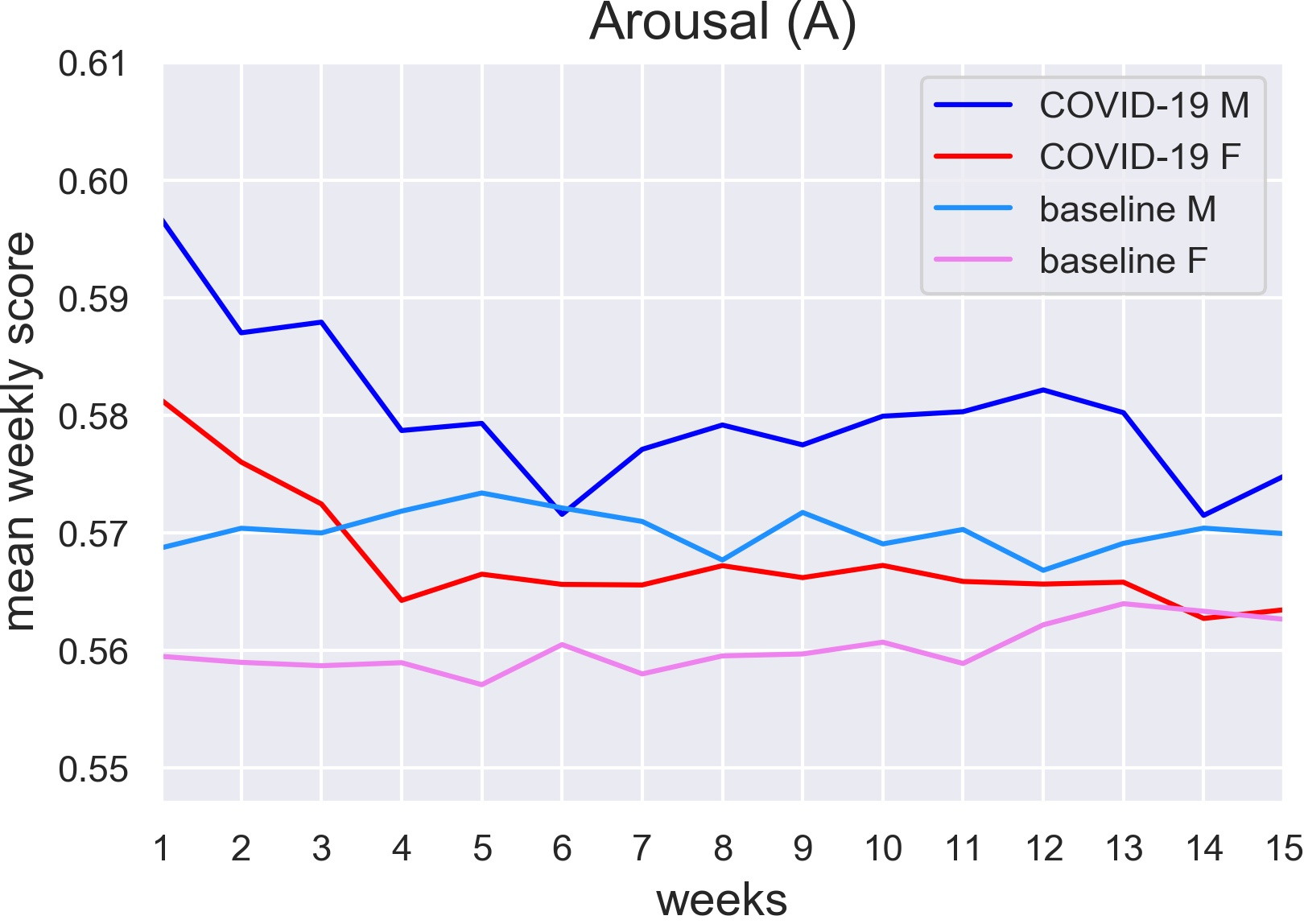}
\end{subfigure}
\qquad \qquad \quad \qquad \qquad \quad
\begin{subfigure}[t]{0.1\textwidth}
    \includegraphics[scale=0.4]{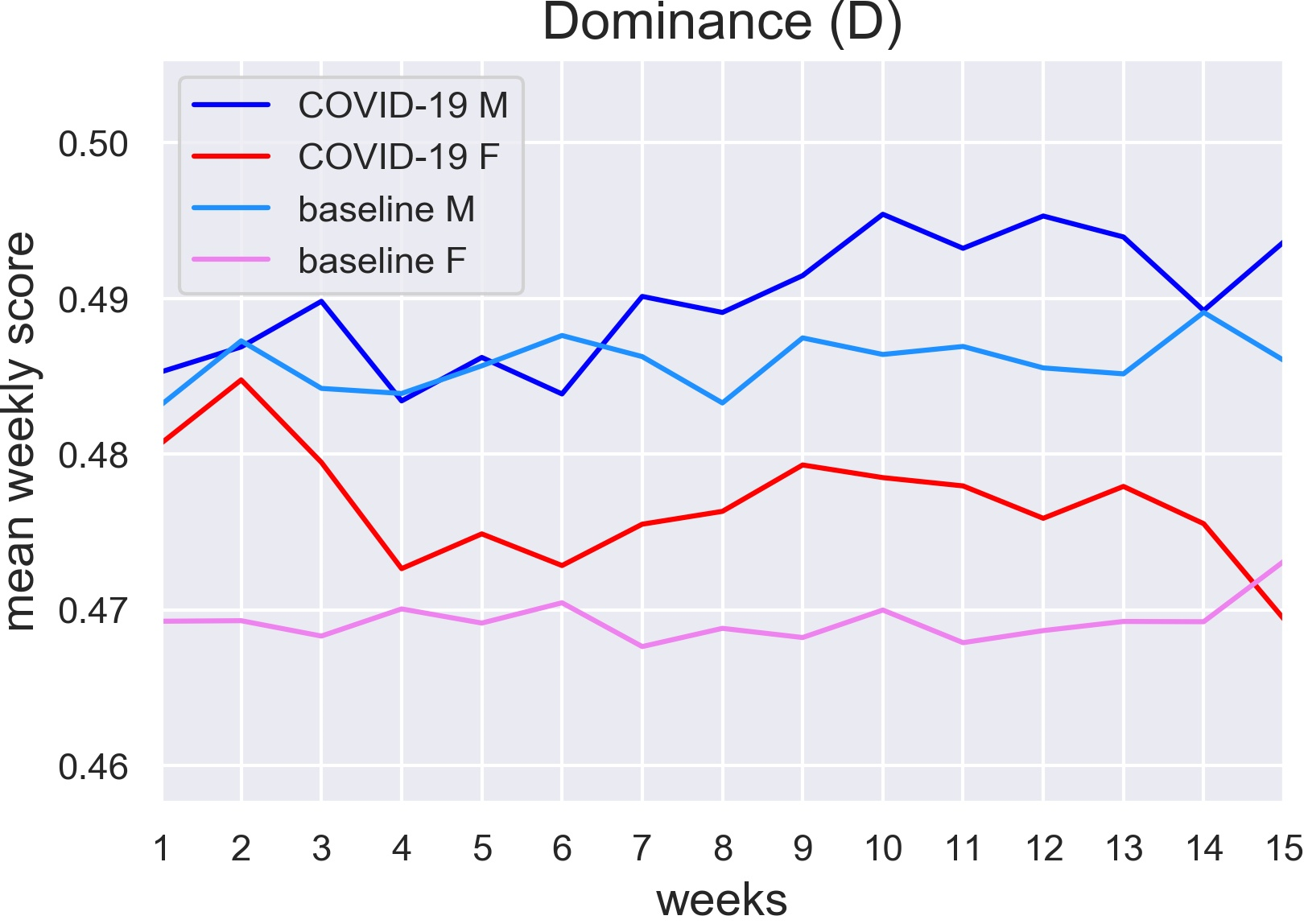}
\end{subfigure}
\caption{\label{fig:vad-diachronic}Diachronic analysis of valence (left), arousal (middle), and dominance (right) scores for Reddit data.}
\end{figure*}

A common way to study emotions in psycholinguistics uses an approach that groups affective states into a few major dimensions, such as the Valence-Arousal-Dominance (VAD) affect representation, where \textit{valence} refers to the degree of positiveness of the affect, \textit{arousal} to the degree of its intensity, and \textit{dominance} represents the level of control \citep{bradley1994measuring}. Computational studies applying this approach to emotion analysis have been relatively scarce due to the limited availability of a comprehensive resource of VAD rankings, with (to the best of our knowledge) no large-scale study on cross-gender language. Here we make use of the recently-released  NRC-VAD Lexicon, a large dataset of human ratings of $20,000$ English words \citep{mohammad2018obtaining}, in which each word is assigned V, A, and D values, each in the range $[0\text{--}1]$. For example, the word `fabulous' is rated high on the valence dimension, while `deceptive' is rated low. 
In this study we aim at estimating the VAD values of posts (typically comprising multiple sentences), rather than individual words; we do so by inferring the affective ratings of sentences using those of individual words, as follows.

Word embedding spaces have been shown to capture variability in emotional dimensions closely corresponding to valence, arousal, and dominance \citep{Hollis2016}, implying that such semantic representations carry over information useful for the task of emotional affect assessment. Therefore, we exploit affective dimension ratings assigned to individual words for supervision in extracting ratings of sentences. We use the model introduced by \citet{ReimersSBERT} for producing word- and sentence-embeddings using Siamese BERT-Networks,\footnote{We used the \texttt{bert-large-nli-mean-tokens} model, obtaining highest scores on a the STS benchmark.} thereby obtaining semantic representations for the $20,000$ words in \citet{mohammad2018obtaining} as well as for sentences in our datasets. This model performs significantly better than alternatives (such as averaging over a sentence's individual word embeddings and using BERT encoding \citep{ReimersSBERT}) on the SentEval toolkit, a popular evaluation toolkit for sentence embeddings \citep{Conneau2018SentEval}.

Next, we trained beta regression models\footnote{An alternative to linear regression in cases where the dependent variable is a proportion (in 0\text{--}1 range).} \citep{zeileis2010beta} to predict VAD scores (dependent variables) of words from their embeddings (independent predictors), yielding Pearson's correlations of $0.85$, $0.78$, and $0.81$ on a $1000$-word held-out set for V, A, and D, respectively. The trained models were then used to infer VAD values for each sentence within a post using the sentence embeddings.\footnote{We excluded sentences shorter than 5 tokens.} A post's final score was computed as the average of the predicted scores for each of its constituent sentences. As an example, the post \textit{`most countries handled the covid-19 situation appropriately'} was assigned a low arousal score of 0.274, whereas a high arousal score of $0.882$ was assigned to \textit{`gonna shoot the virus to death!'}.

\subsection{Results and Discussion}

We compared V, A, and D scores of male posts to those of female posts, in each of the COVID and baseline datasets, using  Wilcoxon rank-sum tests.  All differences were significant, and Cohen's~$d$ \citep{cohen2013statistical} was used to find the  effect size of these differences; see Table~\ref{tbl:vad-values}.  We also compared the scores for each gender in the COVID dataset to their respective scores in the baseline dataset (discussed below). We further show, in Figure~\ref{fig:vad-diachronic}, the diachronic trends in VAD for M and F authors in the two sub-corpora: COVID and baseline.

First, Table~\ref{tbl:vad-values} shows considerable differences between M and F authors in the baseline dataset for all three emotional dimensions (albeit a tiny effect size in valence), in line with established assumptions in this field \citep{burriss2007psychophysiological, hoffman2008empathy, thelwall2010data}: women score higher in use of positive language, while men score higher on arousal and dominance.  Interestingly, the cross-gender differences in V and A are amplified between baseline and COVID data, with an increase in effect size from $0.043$ to $0.120$ for V and $0.109$ to $0.144$ for A.  By comparison, virtually no difference was detected in D between M and F authors in baseline vs.\ virus-related discussions.  Thus we find that men seem to use more negative and emotionally-charged language when discussing COVID than women do -- and to a greater degree than in non-COVID discussion -- presumably indicating a grimmer outlook towards the pandemic.  This finding is particularly interesting, given that \citet{vandervegt2020women} find that women self-report more negative emotion in reaction to the pandemic, and underlies the importance of analysis of implicit indications of affective state in spontaneous text.

COVID-related data trends (Figure~\ref{fig:vad-diachronic}) show comparatively low scores for valence and high scores for arousal in the early weeks of our analysis (February to mid-March). We attribute these findings to an increased level of alarm and uncertainty about the pandemic in its early stages, which gradually attenuated as the population learned more about the virus. As expected, both genders exhibit lower V scores in COVID discussions compared to baseline: Cohen's $d$ effect size of $-0.617$ for M and $-0.554$ for F authors. Smaller, yet considerable, differences between the two sub-corpora exist also for A and D ($0.095$ and $0.047$ for M, and $0.083$ and $0.085$, for F). These affective divergences from baseline show how emotionally-intensive is COVID-related discourse.

\section{Analysis of Topical Distinctions}

\begin{table*}[h!]
\centering
\small
\begin{tabular}{
>{\centering\arraybackslash}p{1.5cm}
>{\centering\arraybackslash}p{1.5cm}
>{\centering\arraybackslash}p{1.5cm}
>{\centering\arraybackslash}p{1.5cm}|
>{\centering\arraybackslash}p{1.5cm}
>{\centering\arraybackslash}p{1.5cm}
>{\centering\arraybackslash}p{1.5cm}
>{\centering\arraybackslash}p{1.5cm} }
\textbf{M-1} & \textbf{M-2} & \textbf{M-3} & \textbf{M-4} & \textbf{F-1} & \textbf{F-2} & \textbf{F-3} & \textbf{F-4}\\
money &     week &      case &      fuck &      virus &     feel &      mask &      week \\
economy &   health &    rate &      mask &      make &      thing &     hand &      test \\
business &  close &     spread &    claim &     good &      good &      wear &      hospital \\
market &    food &      hospital &  news &      thing &     friend &    woman &     sick \\
crisis &    open &      week &      post &      vaccine &   talk &      food &      patient \\ 
make &      travel &    month &     comment &   point &     make &      face &      symptom \\
economic &  supply &    testing &   call &      happen &    love &      call &      doctor \\
pandemic &  store &     social &    article &   human &     parent &    store &     positive \\ 
lose &      stay &      lockdown &  chinese &   body &      anxiety &   close &     start \\
vote &      plan &      measure &   medium &    study &     read &      stay &      care \\
\end{tabular}
\caption{Most coherent topics identified in male (\textbf{M-1}--\textbf{M-4}) and female (\textbf{F-1}--\textbf{F-4}) COVID-related posts.}
\label{tbl:topic-modeling}
\end{table*}

\begin{table*}
\centering
\resizebox{\textwidth}{!}{
\begin{tabular}{l|l|l|c|c}
 &
\multicolumn{1}{c|}{Topic} & \multicolumn{1}{c|}{Keywords} & \multicolumn{1}{c|}{Male} & \multicolumn{1}{c}{Female} \\ \hline
\textbf{1} & \textbf{Economy} & {money, business, make, month, food, economy, market, supply, store, cost}
 & \textbf{0.17} & \textbf{0.10} \\ \hline
\textbf{2} & \textbf{Social} & {feel, thing, live, good, make, friend, talk, love, hard, start}
 & \textbf{0.07} & \textbf{0.26} \\ \hline
3 & Distancing & close, social, health, open, plan, stay, travel, week, continue, risk
 & 0.09 & 0.11 \\ \hline
4 & Virus & virus, kill, human, disease, study, body, spread, effect, similar, immune
 & 0.11 & 0.07 \\ \hline
5 & Health (1) & mask, hand, stop, make, call, good, wear, face, person, woman
 & 0.07 & 0.08 \\ \hline
6 & Health (2) & case, test, hospital, rate, spread, patient, risk, care, sick, testing
 & 0.17 & 0.14 \\ \hline
\textbf{7} & \textbf{Politics} & {problem, issue, change, response, vote, policy, support, power, action, agree}
& \textbf{0.17} & \textbf{0.07} \\ \hline
8 & Media & point, make, question, post, news, read, fact, information, understand, article
 & 0.08 & 0.07 \\ \hline
9 & Misc. & good, start, thing, make, hour, stuff, play, pretty, find, easy
 & 0.08 & 0.10 \\ 
\end{tabular}
}
\caption{\label{tbl:topic-dist} Distribution of dominant topics in the COVID corpus. Entries in columns M(ale) and F(emale) represent the ratio of posts with the topic in that row as their main topic. Ratios are calculated for M and F posts separately (each of columns M and F sum to $1$).  Bolded topics indicate those with substantial differences between M and F.}
\end{table*}

We study topical distinctions in male vs.\ female COVID-related discussions with two complementary analyses: (1) comparison of topics found by topic modelling over each of the M and F subcorpora separately, and (2) comparison of the distribution of dominant topics in M vs.\ F posts as derived from a topic model over the entire M+F dataset.

For each analysis, we used a publicly-available topic modeling tool \citep[MALLET,][]{McCallumMALLET}. Each topic is represented by a probability distribution over the entire vocabulary, where terms more characteristic of a topic are assigned a higher probability.\footnote{Prior to topic modeling we applied a preprocessing step including lemmatization of a post's text and filtering out stopwords (the $300$ most frequent words in the corpus).} A common way to evaluate a topic learned from a set of documents is by computing its \textit{coherence score} -- a measure reflecting 
its overall quality \cite{newman2010automatic}. We assess the quality of a learned model by averaging the scores of its individual topics -- the \textit{model} coherence score. 

\textbf{Analysis of Cross-gender Topics.}
Here we explore topical aspects of the productions of the two genders by comparing two topic models: one created using M posts, and another using F posts, in the COVID dataset.  We selected the optimal number of topics for each set of posts by maximizing its model coherence score, resulting in $8$ topics for male and $7$ topics for female posts (coherence scores of $0.48$ and $0.46$). 

We examined the similarities and the differences across the two topical distributions by extracting the top $4$ topics -- those with the highest individual coherence scores -- in each of the M and F models. Table~\ref{tbl:topic-modeling} presents the $10$ words with highest likelihood for these topics in each model; topics within each are ordered by decreasing coherence score (left to right). We can see that both genders are occupied with health-related issues (topics \textbf{M\text{-}3}, \textbf{F\text{-}1}, \textbf{F\text{-}4}), and the implications on consumption habits (topics \textbf{M\text{-}2}, \textbf{F\text{-}3}). However, clear distinctions in topical preference are also revealed by our analysis: men discuss economy/market and media-related topics (\textbf{M\text{-}1}, \textbf{M\text{-}4}), while women focus more on family and social aspects (\textbf{F\text{-}2}). Collectively these results show that the established postulates regarding gender-linked topical preferences are evident in spontaneous COVID-related discourse on Reddit.

\textbf{Analysis of Dominance of Topics across Genders.}
We next performed a complementary analysis, creating a topic model over the combined male and female sub-corpora, yielding $9$ topics.\footnote{We used the model with the 2nd-best number of topics (9, coherence score 0.432) as inspection revealed it to be more descriptive than the optimal number of topics (2, score 0.450).}  
We calculate, for the two sets of M and F posts, the distribution of dominant topics -- that is, for each of topics $1$--$9$, what proportion of M (respectively F) posts had that topic as its first-ranked topic.


Table~\ref{tbl:topic-dist} reports the results; e.g., row 1 shows that the economy is the main topic of 17\% of male posts, but only 10\% of female posts.  We see that males tend to focus more on economic and political topics than females (rows $1$ and $7$); conversely, females focus far more on social topics than did males (row $2$).  Once again, these findings highlight cross-gender topical distinctions in COVID discussions on Reddit in support of prior results.
\section{Conclusions}
A large body of studies spanning a range of disciplines has suggested (and corroborated) assumptions regarding the differences in linguistic productions of male and female speakers. Using a large dataset of COVID-related utterances by men and women on the Reddit discussion platforms, we show clear distinctions along emotional dimensions between the two genders, and demonstrate that these differences are amplified in emotionally-intensive discourse on the pandemic. Our analysis of topic modeling further highlights distinctions in topical preferences between men and women.

\section*{Acknowledgments}
This research was supported by NSERC grant RGPIN-2017-06506 to Suzanne Stevenson, and by an NSERC USRA to Jai Aggarwal.

\bibliographystyle{acl_natbib}
\bibliography{anthology,main}

\end{document}